\documentclass[runningheads]{llncs}
\usepackage{graphicx}

\begin{document}
\title{
How to reduce computation time while sparing performance during robot navigation? A neuro-inspired architecture for autonomous shifting between model-based and model-free learning \\
}

\titlerunning{ }
%
\author{R{\'e}mi Dromnelle\inst{1}\orcidID{0000-0002-7322-2523} \and
Erwan Renaudo\inst{2}\orcidID{0000-0003-3282-8972} \and
Guillaume Pourcel\inst{1}\orcidID{0000-0002-7147-3652} \and
Raja Chatila\inst{1}\orcidID{0000-0001-7822-0634} \and
Beno{\^i}t Girard\inst{1}\orcidID{0000-0002-3914-6483} \and
Mehdi Khamassi\inst{1}\orcidID{0000-0002-2515-1046}}
\authorrunning{R. Dromnelle et al.}
%
\institute{Sorbonne Universit{\'e}s, CNRS, Institut des Syst{\`e}mes Intelligents et de Robotique (ISIR), F-75005 Paris, France \\
\email{remi.dromnelle@gmail.com} \and
Universit{\"a}t Innsbruck, Intelligent and Interactive Systems Lab (IIS), A-6010 Innsbruck, Austria}
\maketitle              

\begin{abstract}
Taking inspiration from how the brain coordinates multiple learning systems is an appealing strategy to endow robots with more flexibility. One of the expected advantages would be for robots to autonomously switch to the least costly system when its performance is satisfying. However, to our knowledge no study on a real robot has yet shown that the measured computational cost is reduced while performance is maintained with such brain-inspired algorithms. We present navigation experiments involving paths of different lengths to the goal, dead-end, and non-stationarity (i.e., change in goal location and apparition of obstacles). We present a novel arbitration mechanism between learning systems that explicitly measures performance and cost. We find that the robot can adapt to environment changes by switching between learning systems so as to maintain a high performance. Moreover, when the task is stable, the robot also autonomously shifts to the least costly system, which leads to a drastic reduction in computation cost while keeping a high performance. Overall, these results illustrates the interest of using multiple learning systems.

\end{abstract}

\section{Introduction}

The idea of taking inspiration from how the brain coordinates multiple learning systems to enable more flexibility in robots is getting more and more attention in the robotics community \cite{MeyerG2008,dolle2008,CaluwaertsSNGDFGK2012,zambelli2016,banquet2016,lowrey2018}. One of the expected advantages of such a strategy would be for robots to autonomously learn which system is the most appropriate for each encountered task or situation. For instance, a robot can learn that different systems are efficient in different subparts of the environment \cite{CaluwaertsSNGDFGK2012}. Another expected advantage for a robot is to detect when it can avoid the computation time associated to a costly planning process and rely on cheaper systems if they enable to reach the same level of performance.

In computational neuroscience, reinforcement learning (RL) algorithms have been proposed to account for how animals initially solve a new task through planning within a model-based (MB) system, and progressively shift to model-free (MF) control when learning has converged \cite{DawND2005,KhamassiH2012}. MF learning is proposed to represent habit learning because it takes a long time to converge, but permits fast and efficient decisions after learning. Moreover, its slowness in learning makes it inflexible in response to task changes, requiring that the brain switches back to a control level similar to MB.

We have previously proposed a way to implement these principles within a classical three-layered robot cognitive architecture, to facilitate integration with other sensing and control components, as well as permit future transfer to different robotic platforms \cite{RenaudoGCK2015b}. Here, and after evaluating several arbitration mechanisms between MB and MF learning systems in a previous study \cite{RenaudoGCK2015a}, we present a novel one which dynamically deals between the quality of learning and the computation cost. We test the new algorithm during simulated and real robot navigation in a task involving paths of different lengths to the goal, dead-ends, and non-stationarity. We find that the algorithm flexibly and consistently switches to MB control after environmental changes, and to MF control when the task is stationary. Overall, the robot achieves the same performance as optimal MB control in the task, while dividing computation time by more than two. 

In summary, we propose a MB/MF algorithm using an arbitration mechanism that coordinates the learning systems and efficiently reduces computation cost while maintaining performance. We evaluate the algorithm both on simulated and real robots.

\section{Materials and Methods}

\subsection{A robotic architecture with a dual decision-making system}

\begin{figure}[t]
\centering
\framebox{\includegraphics[scale=0.3]{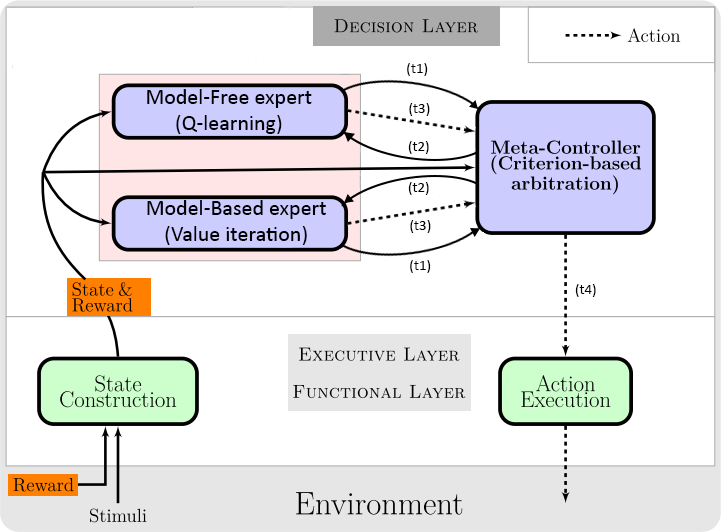}}
\caption{The generic version of the architecture. Two experts having different properties are computing the next action to do in the current state $s$. They each send monitoring data to the meta-controller (MC) about their learning status and inference process (t1). The MC designate the winning expert according to a criterion that uses these data and authorizes it to carry out its inference and decision processes (t2). After making a decision, the winning expert sends its proposition to the MC (t3), which sends the action to the Executive Layer (t4). The effect of the executed action generates a new perception, transformed into an abstract Markovian state, and eventually a non null reward $r$, that are sent to the experts. Each expert learns according to the action chosen by the MC, the new state reached and the reward.}
\label{fig:archi}
\end{figure}

The present work implements a classical three-layer robot cognitive architecture \cite{Gat1998,AlamiCFGI1998} composed of a decision, an executive and a functional layer.
The decision layer of the proposed architecture (Fig. \ref{fig:archi}) is composed by two competing experts which generate action propositions, each with its own method and with its own advantages and disadvantages. These two experts are directly inspired by the currently conventional distinction in computational neuroscience models between goal-directed and habitual strategies \cite{KhamassiH2012}. The two experts run three processes in a row: learning, inference and decision. This layer is also provided with a meta-controller (MC) in charge of arbitrating between experts. The MC determines which expert's proposed action will be executed in the current state, according to an arbitration criterion.

After that, the decision layer sends the chosen action to the executive layer, who ensures its accomplishment by recruiting robot's skills from the functional layer. The latter consists of a set of reactive sensorimotor loops that control actuators during interaction with the environment. The robot reaches a new state and obtains or not a reward. The two experts use the new state and the reward information to update their knowledge about the executed action. This allows MB and MF experts to cooperate by learning from each others' decision. 

Compared to our previous architecture \cite{RenaudoGCK2015a}, several changes have been made: The overall organization of the decision-making layer and the prioritization of communication between modules have been changed. The MF expert is no longer built as a neural network but as a tabular algorithm. The MC chooses which expert is the most suitable at a given time and in a given state, and no longer simply at a given time. And above all, we have defined a novel arbitration criterion that allows to reduce computational cost while maintaining performance.

\subsection{The decision layer}

\subsubsection{Model-based expert.}

The MB expert learns a transition model $T$ and a reward model $R$ of the problem, and uses them to compute the values of actions in each state. These models allow to simulate over several steps the consequences of following a given behavior and to look for desirable states to reach. Consequently, when the robot realizes that the task has changed, it can use this knowledge of the world to instantly find the new relevant behavior. However, this search process is costly in terms of computation time as it needs to simulate several value iterations \cite{SuttonB1998} in each state to find the correct solution.

\paragraph{Learning process.}

The learning process of the MB consists in updating the reward and the transition models by interacting with the world. The transition model $T$ is learnt by counting occurrences of transitions $(s,a,s')$. We build it using the number of visits $V_{N}(s,a)$ of state $s$ and action $a$. $V_{N}(s,a)$ has a maximum value of $N$ and $V_{N}(s,a,s')$ is the number of visits of the transition $(s,a,s')$ in the last $N$ visits of $(s,a)$. The transition probability $T(s,a,s')$ is defined in (1). This leads to an estimation of the probability to the closest multiple of $1/N$.

$$
T(s,a,s') = \frac{V_{N}(s,a,s')}{V_{N}(s,a)}
\eqno{(1)}
$$

The reward model $R$ stores the most recent reward value $r_t$ received for performing action $a$ in state $s$ and reaching the current state $s'$, multiplied by the probability of the transition (s,a,s').

\paragraph{Inference process.}

Performing the process of inference consists in planning using a tabular Value Iteration algorithm \cite{SuttonB1998}: 

$$
Q(s,a) \leftarrow \sum_{s'} T(s,a,s') \left[ R(s,a) + \gamma max_{a'}Q(s',a') \right]
\eqno{(2)}
$$

$Q(s,a)$ is the action-value estimated by the agent for performing the action $a$ in the state $s$, $R(s,a)$ the probabilistic reward of the reward model $R$ associated with the transition $(s,a)$ and $\gamma$ the decay rate of future rewards.


\paragraph{Decision process.}

Performing the decision process consists in converting the estimation of action-values into a distribution of action probabilities using a softmax function, and drawing the action proposal from this distribution:

$$
P(a|s) = \frac{ \exp( Q(s,a) / \tau )  }{ \sum_{b \in \mathcal{A}} \exp(Q(s,b) / \tau )}
\eqno{(3)}
$$

\noindent where $\tau$ is the exploration/exploitation trade-off parameter.

\subsubsection{Model-free expert.}

The MF algorithm does not use models of the problem to decide which action to do in each state, but directly learns the state-action associations by caching in each state the earned rewards in the value of each action (action-values). Because updating the action-values is local to the visited state, the process is slow and the robot cannot learn the topological relationships between states. Consequently, when the task changes, the robot takes many actions to adopt the new relevant behavior. On the other hand, this method is less expensive in terms of inference duration.

\paragraph{Learning process.}

Performing the learning process consists in estimating the action-value $Q(s,a)$ using a tabular Q-learning algorithm:

$$
Q(s,a) = Q(s,a) + \alpha \left[ R(s) + \gamma max_{a'}Q(s',a') - Q(s,a) \right]
\eqno{(4)}
$$

$R(s)$ is the instant reward received for reaching the state $s$ and $\gamma$ the decay rate of future rewards and the $s'$ the state reached after executing $a$.

\paragraph{Inference process.}

Since the MF expert does not use planning, its inference process consists only in reading from the table that contains all the action-values the one that corresponds to performing the action $a$ in the state $s$.

\paragraph{Decision process.}

The decision process is the same as for the MB expert (3). 

\subsubsection{\label{mcorga}Meta-controller and arbitration method.}

The MC is in charge of selecting which expert will generate the behavior using a novel arbitration criterion which is a trade-off between the quality of learning and the cost of inference. 

\paragraph{Quality of learning.}

At each time step $t$, if the expert $E$ is selected to lead the decision, its action selection probabilities (3) are filtered using a low-pass filter and stored by the system :

$$
f(P(a|s,E,t)) = (1.0 - \alpha) * f(P(a|s,E,t-1)) + \alpha * P(a|s,E,t)
\eqno{(5)}
$$

\noindent Else, no low-pass filter is applied :

$$
f(P(a|s,E,t)) =  f(P(a|s,E,t-1))
$$

Using the filtered action probability distribution $f(P(a|s,E,t))$, the MC can compute the entropy $H(s,E,t)$ of each expert, which has previously been found to reflect the quality of learning in humans \cite{ViejoKBG2015} :

$$
H(s,E,t) = - \sum_{a=0}^{|\mathcal{A}|} f(P(a|s,E,t)) \cdot{} log_2(f(P(a|s,E,t)))
\eqno{(6)}
$$

\paragraph{Cost of inference.}

At each time step $t$ and for each state $s$, the duration $T(s,E,t)$ of the inference process of each expert is recorded and filtered in the same way as the action selection probabilities (5).

\paragraph{Arbitration criterion.}

Using the quality of learning and the cost of inference, the MC computes one expert-value $Q(s,E,t)$ for each expert: 

$$
Q(s,E,t) = - (H(s,E,t) + \kappa T(s,E,t))
\eqno{(7)}
$$

\noindent where $\kappa = e^{-\eta H(s,MF,t)}$ allows to weight the impact of time in the criterion: The lower the entropy of the distribution of MF action probabilities, the more weight the time taken to perform the inference process has in the equation. $\eta$ is a constant parameter ($\eta=7$) weighting the entropy term and set according to a Pareto front analysis \cite{powel2015} (not shown here). We were looking for a $\kappa$ that minimizes the cost of inference, while maximizing the agent's ability to accumulate reward over time. 

Finally, the MC converts the estimation of expert-values $Q(s,E,t)$ into a distribution of expert probabilities using a softmax function (3), and draws the expert proposal from this distribution. The inference process of the unchosen expert is inhibited, which thus allows the system to save computation time.

\subsubsection{General information.}

Similarly to the Rmax algorithm \cite{SuttonB1998}, we initialized the action-values to a value of 1 so to help exploration of non-previously selected actions, since the action-values are updated according to the previous ones. 

For the MF expert, we conducted a grid search to find the best parameter-set, i.e. parameters maximizing the total accumulated reward over a fixed duration of 1600 timesteps. We experimentally found that the duration of 1600 actions is a good trade-off between the time needed by the MF to start learning and a reasonable experiment time (1600 actions correspond to about 5 hours of real experiment). We found $\alpha$ = 0.6, $\gamma$ = 0.9 and $\tau$ = 0.02. For the MB expert, we chose $\gamma$ = 0.95 and $N$ = 6. For the MB expert and the MC, we chose the same value of $\tau$ as the MF expert.

\subsection{The experimental task}

We evaluated our cognitive architecture in a navigation task. Since running 1600 actions on the robot takes about six hours, we have created a simulation of the task where the probabilities of transitions are derived from a 13 hours exploration of the real arena (without the reward). This simulation allowed us to quickly test multiple coordination criteria and parameterizations, before evaluating them on a real robot. 

\begin{figure}[thpb]
\centering
\framebox{\includegraphics[scale=0.066]{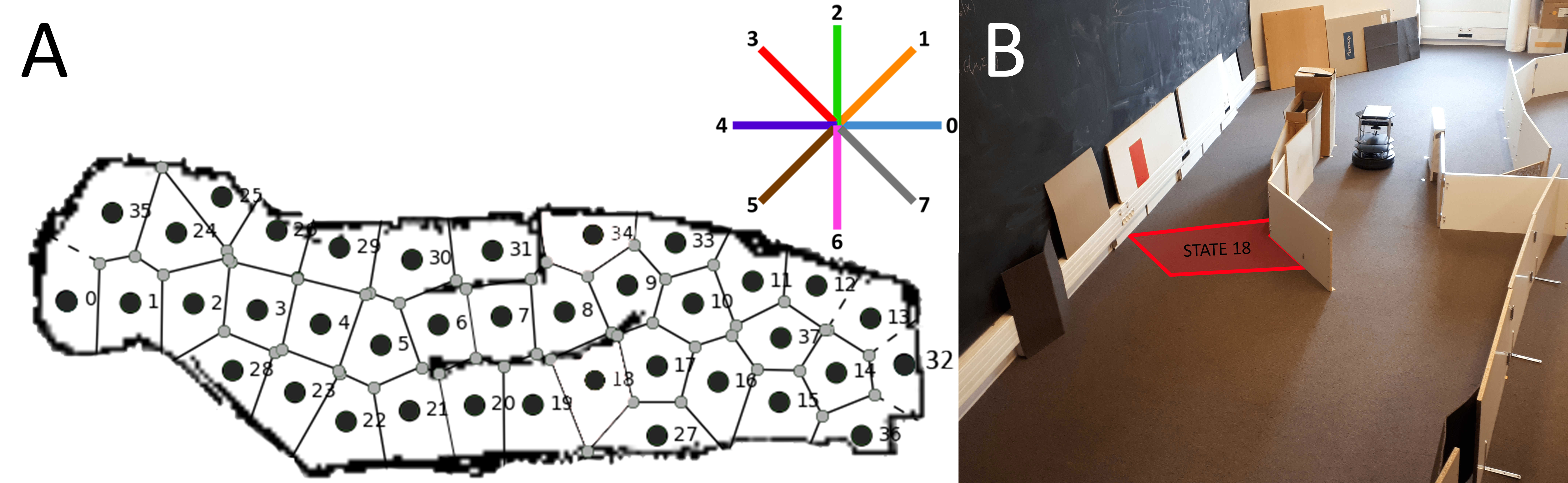}}
\caption{\textbf{A}. Map of the arena's states. The eight-pointed star indicates the direction (in the map) of each robot actions. \textbf{B}. Photo of the arena and a turtlebot heading into the middle corridor. The state 18 (initial reward location) is represented in red.}
\label{fig:map_actions_photo}
\end{figure}

We used a 2.6 m x 9.5 m arena containing obstacles (Fig \ref{fig:map_actions_photo}), and a turtlebot.The computer uses ROS \cite{QuigleyCGFFLWN2009} to process the signals from its sensors, controls the mobile base and interfaces with our architecture. A Kinect-1 sensor returns an estimate of distance to obstacles in its field of view, completed by contact sensors at the front and sides of the mobile base. The robot localizes itself using the gmapping Simultaneous Location and Mapping Algorithm (SLAM, \cite{GrisettiSB2007}). During a preliminary environmental exploration phase, the robot incrementally builds a topological map by adding evenly spaced centers, and thus autonomously creating new Markovian states (Fig. \ref{fig:map_actions_photo}.A). The current state (of the corresponding MDP) is the closest center from the robot when its previous action is completed and it evaluates the consequences. We chose to build this map beforehand and to reuse it for each of the learning experiments, so as to reduce the sources of behavioral variability. However, note that with the present method the system could start with an empty map and build it incrementally, and that a new map could be used for each experiment.

In this experiment, the robot must learn to reach a specific state of the environment (state 18 -- see Fig. \ref{fig:map_actions_photo}.B). When it succeeds, it receives a unitary reward and is randomly returned to one of the two initial positions, located in the extremities of the arena (states 0 and 32), to start over. The goal of the robot is first to reach state 18. The experiment involves a stable period where the environment and reward do not change (i.e., until action 1600), followed by a task change where the reward is moved from state 18 to state 34. We also made a second series of experiments where the reward is fixed but obstacles are introduced in the environment. As the state space, the action space is a discrete space. Here, performing an action consists of going forward along 8 equally distributed allocentric directions (Fig. \ref{fig:map_actions_photo}.A). As long as the robot has not changed state, the action is not considered as completed. However, if while the robot moves forward, its contact sensors are activated (it bumps into a wall), then it will move back 0.15 meters and the action is considered as completed. Finally, according to the exact position in which the agent is located within a state, the arrival state will not necessarily be identical for the same action performed. The environment is therefore probabilistic, which multiplies the possibilities for the agent. For the MB expert, this specificity implies that the transitions $T(s,a,s')$ and the rewards $R(s,a)$ are stored respectively in the model of transition $T$ and the model of reward $R$ as probability distributions. 

\section{Results}

We first present the results obtained when a virtual agent performs the task in a simulated environment, and then, the replication of these results in the real environment with a Turtlebot.

\begin{figure}[t]
\centering
\framebox{\includegraphics[scale=0.1345]{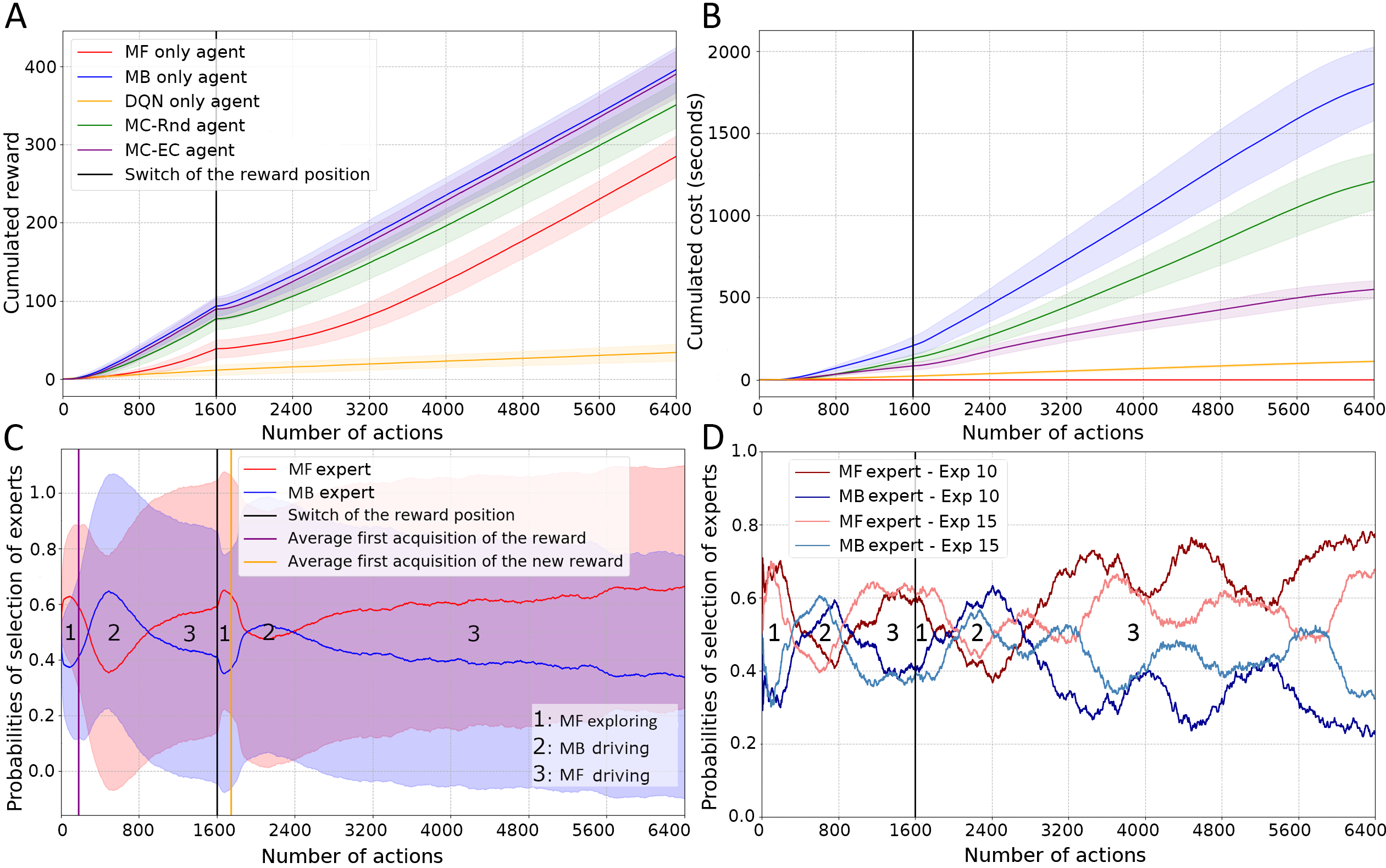}}
\caption{\textbf{A}. Mean performance for 100 simulated runs of the task. The performance is measured as the cumulative reward obtained over the duration of the experiment. The duration is represented as the number of actions performed by the agent. We use standard deviation as dispersion indicator. At the 1600th action, the reward switches from the state 18 to the state 34. \textbf{B}. Mean computational cost for 100 simulated runs of the task. The computational cost is measured as the cumulative time of the inference process over the duration of the experiment in seconds. \textbf{C}. Mean probabilities of selection of experts by the MC using the Entropy and Cost criterion for 100 simulated runs of the task. These probabilities are defined by the softmax function of each expert. \textbf{D}. Probabilities of selection of experts by the MC using the Entropy and Cost criterion for 2 simulated runs of the task. }
\label{fig:CR_CT_PS_sim_SR}
\end{figure}

\subsection{Simulated task}

To evaluate the performance of the virtual agent, we studied four combinations of experts : (1) a MF only agent using only the MF expert to decide, (2) an MB only agent using only the MB expert to decide, (3) a random coordination agent (MC-Rnd) which coordinates the two experts randomly and (4) an Entropy and Cost agent (MC-EC) which coordinates the two experts using the model of arbitration presented in \ref{mcorga}. We also compare our agent to an agent using a reference learning algorithm in the literature, a DQN \cite{mnih2015}. We evaluated iteratively several networks with various number of layers and size of layers, and selected the set of parameters that achieved the best performance. The neural network composed of two hidden layers of 76 neurons which takes as input a vector of size 38 (corresponding to the activity of the states, with 1 if the state is active, and 0 if not), returns a vector of size 8 (corresponding to the 8 action-values of the active state) and uses experience replay. Its parameters are $\alpha$ = 0.1 $\gamma$ = 0.95 and $\tau$ = 0.05.  

We define the "optimal behaviour" as the behaviour that allows the agent to accumulate the most reward over time (Fig. \ref{fig:CR_CT_PS_sim_SR}.A). As expected, the MF only agent (red) takes longer to reach the optimal behaviour. On the other hand, the MB only agent (blue) has the best performance. The MC-EC agent (purple) has a non-significantly different performance from the MB only agent, showing that our coordination method does not penalize the agent in terms of cumulated reward. In addition to that, it performs better than the MC-Rnd agent (green) suggesting that our coordination method is more effective than chance to accumulate reward over time. At the 1600th action, the environment is modified (change of reward state). The MF only agent takes longer to recover from environmental change than the other agents. Indeed, the MF expert does not use planning method and only updates its action-values locally: a method that takes longer to be effective. Finally, we can observe that the DQN agent learns and adapts less well than all other agents. As it is a model-free algorithm, it is not surprising that agents using the MB expert are more efficient and adaptive. The DQN is also worse than our tabular MF because it has much more memorized values (i.e. the weights of the network) to adapt before being able to provide correct outputs: the training of deep neural networks generally require several hundred thousand iterations. Such number are much too large, when targeting applications to real robot experiments, where learning on-the-fly is required. Replay mechanisms, or training in simulation, could be used to speed-up learning of the DQN, but these additional computations would clearly increase the computational cost of the resulting system.

Unsurprisingly, the MF only agent has a very low computational cumulated cost (Fig. \ref{fig:CR_CT_PS_sim_SR}.B) since its inference process simply consists in reading from the table that contains all the actions-values, while the MB only agent has a high computational cost, because its inference process is a planning method. The MC-EC agent, which exhibits a performance similar to the MB, has a computational cost three times smaller: the average cumulative time at the end of the experiment spent by the MB only agent on its inference process is 1750s versus 500s for the MC-EC agent at action 6400. It is to be noted that the meta-controller has in any case a very low cost, similar to the MF expert, of $10^-5$ seconds per iteration on average. In this system, only the MB expert is expensive, with an average cost of $10^-2$ seconds per iteration. The cost of using a meta-controller is therefore negligible compared to what it brings in terms of overall savings. 

The dynamics of the selection of the experts by the MC, expressed in terms of selection probabilities (Fig. \ref{fig:CR_CT_PS_sim_SR}.C), displays three different phases:

\textbf{The MF exploring phase (1 on Fig. \ref{fig:CR_CT_PS_sim_SR}.C).} Before the discovery of the position of the reward, the agent uses mainly the MF expert. This is due to the difference in the method for updating action-values between the two experts. With the same initial values and the set of parameters we have defined, the action-values of the MF expert decrease slightly more than those of the MB expert, which drives a more pronounced decrease of the entropy of the action probability distribution. In addition, since we do not have an expert specialized in exploration, it makes sense to use the cheapest expert until the position of the reward has been discovered. About exploration, other studies propose to deal between three experts: a MB expert, a MF expert and an expert specialized in the exploration of the environment \cite{CaluwaertsSNGDFGK2012}.

\textbf{The MB driving phase (2 on Fig. \ref{fig:CR_CT_PS_sim_SR}.C).} After finding the first reward the MB expert progressively takes the lead on the decision because its process of inference needs only to find the reward once to spread action-values into its transition model. It finds the reward more easily than the MF expert, and so, its performance increases.

\textbf{The MF driving phase (3 on Fig. \ref{fig:CR_CT_PS_sim_SR}.C).} The MF expert learns by demonstration from the MB expert, and thus spreads action-values from state to state and eventually, towards the 800th action, it reaches the performance of the MB expert. Because the MF expert is less expensive, the model of arbitration gives it the lead on the decision.

A MF exploring phase starts again at the 1600th action when the rewarded state moves from state 18 to 34. Then, the MB driving and the MF driving phases repeat. 


The large standard deviation is explained by the fact that for each experiment, the agent's strategy and behaviour can be very different, notably due to the large number of states and possible actions, but also to the probabilistic nature of the environment. As a result, the time of the switches from one phase to another varied a lot from one individual to another. Nevertheless the individual behavior of each run is consistent with the average behavior presented here (Fig. \ref{fig:CR_CT_PS_sim_SR}.D).

\subsection{Real task}

\begin{figure}[h]
\centering
\framebox{\includegraphics[scale=0.0675]{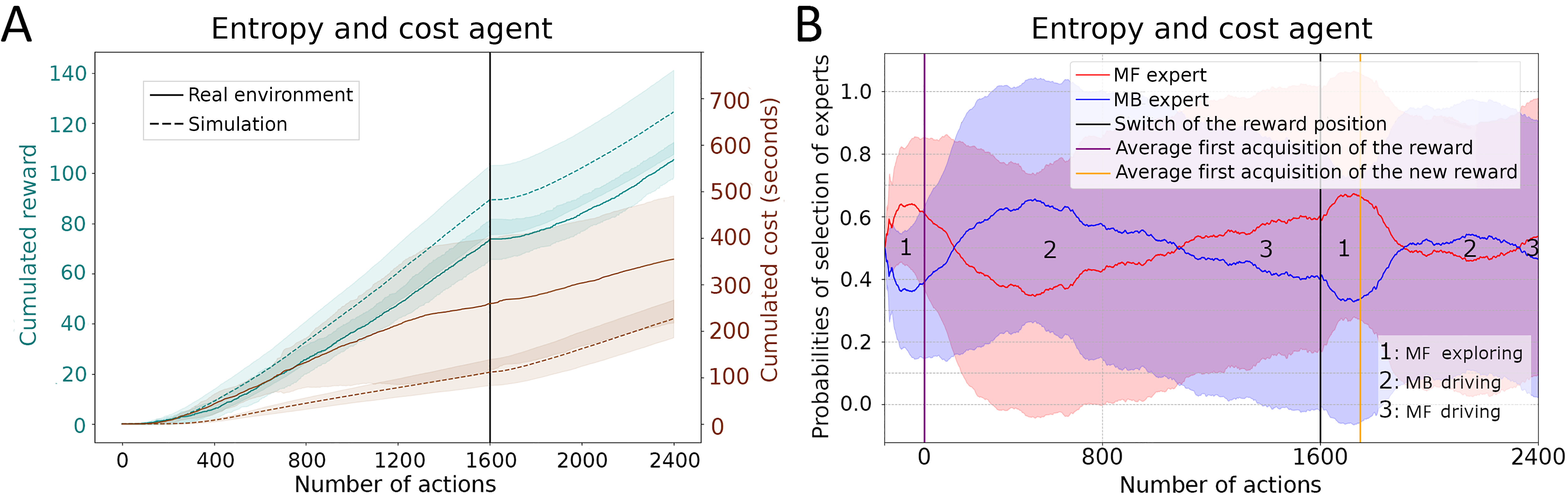}}
\caption{\textbf{A}. Mean performance (brown) and cost (cyan) for 100 simulated runs (dashed curves) and 10 real runs (solid curves) of the navigation task for the MC-EC agent. \textbf{B}. Mean probabilities of selection of experts by the MC using the Entropy and Cost criterion for 10 real runs of the task.}
\label{fig:CR_CT_PS_real_SR}
\end{figure}

We evaluated our model of coordination on a real robot to verify that these results cross the reality-gap. Fig. \ref{fig:CR_CT_PS_real_SR}.A compares the performance and the cost of the MC-EC agent and the real robot (both use the same model of arbitration). The reality gap is visible, with a drop in performance and a cost increase for the real robot compared to the simulation. However, the model still allows the real robot to learn and accumulate reward over time in the same way, and the economy of cost remains advantageous.

Fig. \ref{fig:CR_CT_PS_real_SR}.B shows the dynamics of selection of the experts by the MC, for the experiments in real environment with the real robot. Again, the three-phases pattern is present, with only a 300 actions mean delay at the beginning of the third phase.

We obtained similar strategy alternations with the environment change consisting of obstacles introduction without moving the reward. We also observed that geographical patterns of coordination of experts emerged over time. These results won't be presented in details here because of space limitations.

\section{Discussion}

We analyzed the behavior of a three-layered robotic architecture integrating neuro-inspired mechanisms for the coordination of MB and MF reinforcement learning. The novelty relies in the explicit online measure of performance and cost of each system, so as to give control to the system with best current trade-off between the two. We presented real and simulated navigation results in a complex, non-stationary indoor environment. The arbitration criterion proposed in this work allowed the robot to autonomously determine when to shift between systems during learning, generating a coherent temporal decision-making pattern that alternates between strategies over time. This promoted more flexibility than pure MF control in response to task changes, and permitted to reach the same level of performance than pure MB control, while dividing computation time by three.
The comparison with DQN showed that using end-to-end RL has a computational cost not compatible with robotic constraints, and that thus building and using a data representation adapted to the task at hand reduces the burden on the RL part of the system, allowing for low-cost on-the-fly learning.
In future work, we plan to test whether this architecture is generalizable to other scenarios and larger spaces states, which we have already begun to do by applying our model to a social interaction task defined by 112 states. \cite{dromnelleROMAN2020}.


\bibliographystyle{ieeetr}
\bibliography{article}

\end{document}